\documentclass[conference]{IEEEtran}
\IEEEoverridecommandlockouts
\usepackage{cite}
\usepackage{amsmath,amssymb,amsfonts}
\usepackage{graphicx}
\usepackage{textcomp}
\usepackage{xcolor}
\usepackage{multirow}
\usepackage{subfigure}
\usepackage{caption}
\usepackage{amsfonts}
\usepackage{verbatim}
\usepackage{graphicx}
\usepackage{amsmath}
\usepackage{adjustbox}
\usepackage{subfigure}
\usepackage{footmisc}
\usepackage{booktabs}
\usepackage{algorithm2e}
\def\BibTeX{{\rm B\kern-.05em{\sc i\kern-.025em b}\kern-.08em
    T\kern-.1667em\lower.7ex\hbox{E}\kern-.125emX}}
\begin{document}

\title{RDIS: Random Drop Imputation with Self-Training for Incomplete Time Series Data}
\author{\IEEEauthorblockN{Tae-Min Choi$^\ast$\thanks{$^\ast$ Equal Contribution.}}
\IEEEauthorblockA{School of Electrical Engineering \\
\textit{KAIST}\\
Daejeon, South Korea \\
tmchoi@rit.kaist.ac.kr}
\and
\IEEEauthorblockN{Ji-Su Kang$^\ast$}
\IEEEauthorblockA{KLleon R\&D Center \\
Seoul, South Korea \\
jisukang618@gmail.com}
\and
\IEEEauthorblockN{Jong-Hwan Kim}
\IEEEauthorblockA{School of Electrical Engineering\\
\textit{KAIST}\\
Daejeon, South Korea \\
johkim@rit.kaist.ac.kr}}

\maketitle

\begin{abstract}
Time-series data with missing values are commonly encountered in many fields, such as healthcare, meteorology, and robotics. The imputation aims to fill the missing values with valid values. Most imputation methods trained the models implicitly because missing values have no ground truth. In this paper, we propose Random Drop Imputation with Self-training (RDIS), a novel training method for time-series data imputation models. In RDIS, we generate extra missing values by applying a random drop on the observed values in incomplete data. We can explicitly train the imputation models by filling in the randomly dropped values. In addition, we adopt self-training with pseudo values to exploit the original missing values. To improve the quality of pseudo values, we set the threshold and filter them by calculating the entropy. To verify the effectiveness of RDIS on the time series imputation, we test RDIS to various imputation models and achieve competitive results on two real-world datasets.
\end{abstract}


\section{Introduction}

Due to technological advancements in sensors and other hardware, a massive amount of time-series data can be collected, which can be used to provide services in various environments such as IoT and autonomous driving \cite{batres2015deep, bauer2016arrow, iglesias2013analysis, hayrinen2008definition, tang2020joint}. However, data collection errors in time-series data often occur due to sensor malfunction and human mistakes \cite{silva2012predicting}. Incomplete data means partially lost data; thus, incomplete data comprises missing and observed values. One of the ways to deal with incomplete data is the imputation to fill in the missing values.

The goal of time series imputation is to fill in the missing values in the incomplete time series data. Most imputation works are challenging to train the model explicitly because missing values have no ground truth. Nevertheless, various imputation methods using deep learning \cite{yi2016st, yoon2018gain, luo20192} have recently been developed. They used the time lag between observed values as an additional input and proposed a novel structure specialized for imputation \cite{che2018recurrent, ma2019end, yoon2017multi, cao2018brits, luo2018multivariate}. However, they had to rely on the reconstruction loss of observed values for model training due to the ground truth. The reconstruction loss of observed values helps the learning stability of the imputation model, but it is not directly learning about imputation. To this end, we randomly drop the observed values in the incomplete time series data and train the model by imputing these values. 

Self-training \cite{tarvainen2017mean, berthelot2019mixmatch, zou2019confidence, xie2020self} is one of the methods in semi-supervised learning (SSL), which generates pseudo labels as ground truth of unlabeled data using pre-trained models trained with labeled data. Self-training can also be used for imputation by treating observed and missing values with labeled and unlabeled data. It is essential to use high-quality pseudo labels for proper self-training. In general, we can evaluate the reliability of pseudo labels with deterministic probabilities in the image classification \cite{xie2019self, yalniz2019billion} and the semantic segmentation \cite{bellver2019budget}. However, imputation is not a deterministic task; it is not easy to measure the reliability of the pseudo label generated from the pre-trained model. In this light, we generate several imputed values for one missing value using multiple models and evaluate the confidence of the pseudo label with their entropy.

In this paper, we propose a novel training method for data imputation called random drop imputation with self-training (RDIS). RDIS is composed of two parts: random drop imputation (RDI) and self-training. RDI randomly removes the observed values in the incomplete data and trains an imputation model to impute these values. Thus, unlike the previous methods, which reconstruct observed values in original data in the training process, RDI explicitly trains an imputation model. Furthermore, numerous augmented data can be generated by the random drop. To utilize these augmented data, RDI can employ ensemble learning.

Although incomplete data consists of missing and observed values, RDI only utilizes the observed values to train the imputation model because missing values do not have ground truth. To deal with this issue, RDIS is designed, which adds self-training to RDI to train the imputation model utilizing the missing values in the original incomplete data. Self-training is part of SSL using pseudo labels generated from a teacher (pre-trained) model on unlabeled data. Self-training with the pseudo label is usually used in deterministic tasks, but we propose \textit{pseudo values} instead of pseudo labels since we aim to generate continuous values. RDIS has three main steps: 1) train the teacher (pre-trained) models using RDI, 2) use the teachers to generate pseudo values on missing values, and 3) train each model on the combination of observed values and pseudo values. Then, not every pseudo value is reliable, so we calculate the entropy of the pseudo values to measure the confidence of each pseudo value. To the best of our knowledge, this paper makes the first attempt to adopt self-training in a task of imputation.

In summary, the contributions of this paper are three-fold:
\begin{itemize}
\item We propose RDI to train imputation explicitly. While the previous methods learn imputation by reconstructing the observed values in the incomplete data, RDI trains imputation directly. Since RDI randomly removes the observed values in the incomplete data, numerous augmented data can be generated. Therefore, we adopt ensemble learning by training multiple models with such augmented data.
\item We propose RDIS, which combines self-training with RDI to exploit missing values in incomplete data by generating pseudo values. To measure the reliability of pseudo values, we calculate the entropy by utilizing the ensemble model pre-trained by RDI.
\item We evaluate our method on two multivariate real-world datasets: air quality and gas sensor datasets. Experimental results show that our method performs better than the previous models for imputation accuracy.
\end{itemize}

\section{Related Works}

Since imputation methods using machine learning \cite{nelwamondo2007missing, hudak2008nearest, royston2011multiple, moritz2017imputets} have surged and shown its great advantages, deep learning implementation has been studied in imputation tasks, which have recently achieved more developments. In particular, in the case of time-series data, imputation methods using the RNNs \cite{che2018recurrent, cao2018brits, ma2019end} have gained much attention. \cite{che2018recurrent} developed GRU-D, adding a decay term to the GRU cell so that the model can learn the effect of the temporal gap between missing data. \cite{cao2018brits} used bidirectional RNN to improve imputation performance. Inspired by the residual network (ResNet) \cite{he2016deep}, \cite{ma2019end} added a weighted linear memory vector to RNN to make the model less affected by missing data. Instead of learning imputation directly, indirect learning was rather used in deep-learning-based studies by training a model with only the observed values in incomplete data \cite{che2018recurrent, cao2018brits} or training with classification loss \cite{ma2019end}. 

Imputation methods using generative models \cite{yoon2018gain, luo2018multivariate, luo20192, liu2019naomi, tashiro2021csdi} also have made significant advances. GAN \cite{goodfellow2014generative} guides a model to learn the distribution of the original data to generate data with a similar distribution, allowing the model to learn imputation. \cite{yoon2018gain} adopted conditional GAN by using the original data and mask as the condition. \cite{luo2018multivariate} designed a recurrent network using GRU-I, a simplified version of GRU-D, and applied GAN. Similarly, \cite{luo20192} used GRU-I, but the difference is that it designed a recurrent autoencoder. Unlike the existing autoregressive methods, \cite{liu2019naomi} developed a method training in a non-autoregressive fashion. \cite{liu2019naomi} also uses the divide and conquer strategy to fill in the missing values gradually rather than filling it at once. \cite{tashiro2021csdi} adopted a conditional score-based diffusion model \cite{ho2020denoising} for imputation. The conditional diffusion model is explicitly trained for imputation and exploits correlations between observed values.

Semi-supervised learning (SSL) is a class of algorithms to train with both labeled and unlabeled data \cite{lee2013pseudo, laine2016temporal, tarvainen2017mean, berthelot2019mixmatch}. One of the popular approaches for SSL is self-training, which uses an ensemble model or a teacher model to generate pseudo labels for the unlabeled data. Recently, self-training has been introduced in various fields, such as image classification \cite{xie2019self, yalniz2019billion}, semantic segmentation \cite{bellver2019budget}, unsupervised domain adaptation \cite{zou2018unsupervised, deng2019cluster}. In the task of imputation, we can treat missing values as unlabeled data and observed values as labeled data since missing values do not have ground truth. In this light, we adopt self-training for the imputation task.

\begin{figure*}[t]
  \centering 
  \includegraphics[width=\linewidth]{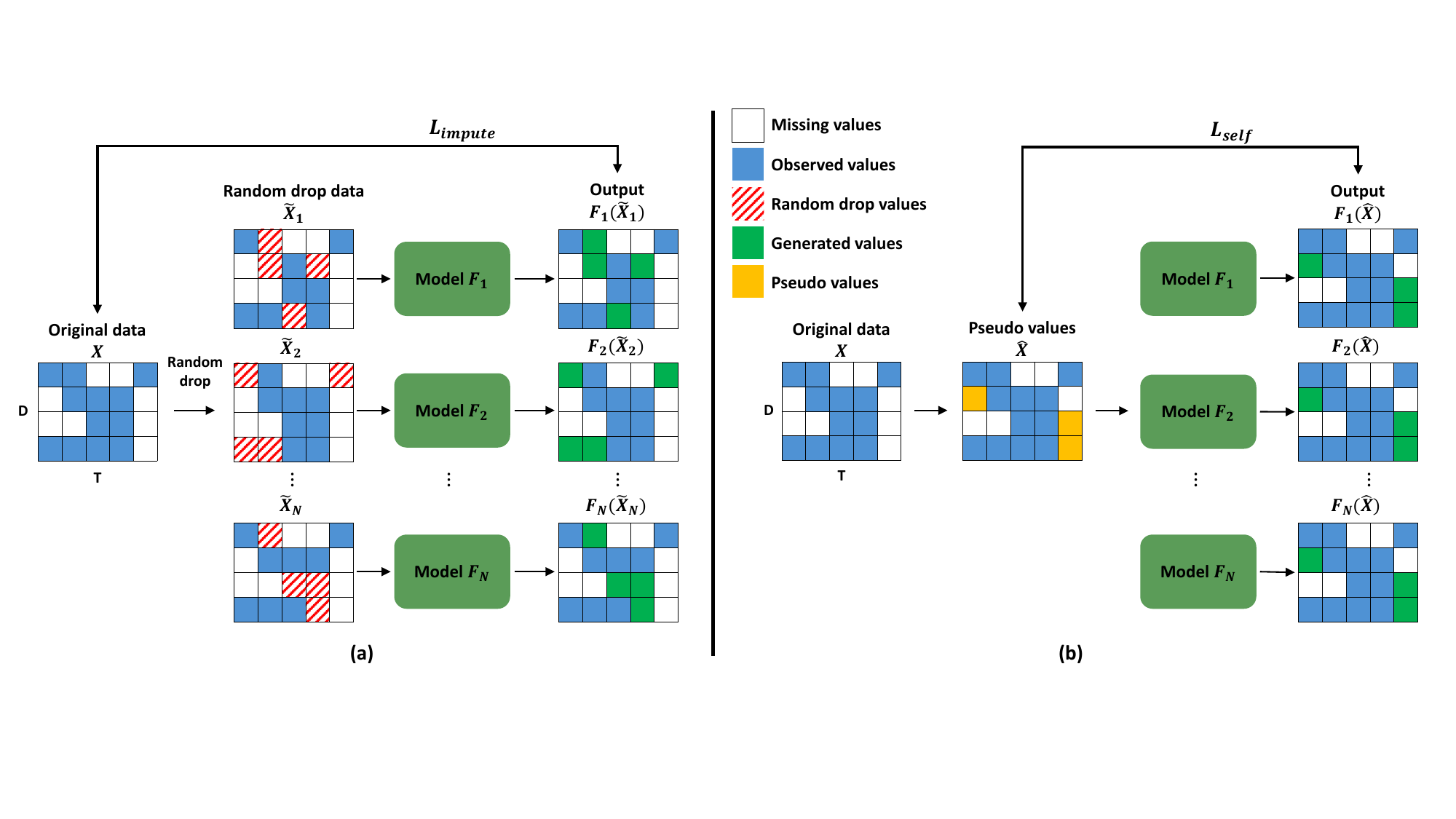}
  \caption{(a) The training procedure of the proposed RDI. The white, blue, and green areas represent missing, observed, and generated values. We generate random drop data represented by a red slashed box by dropping some observed values. Original data can be augmented to $N$ random drop data. Then, we feed each random drop data to each model and calculate $L_{impute}$ using the original data and the output. (b) The training procedure of RDIS. The yellow areas represent pseudo values with enough confidence higher than the threshold. We feed original data with pseudo values and calculate $L_{self}$ using the pseudo values and the output.}
\end{figure*}

\section{Method}
RDIS consists of RDI and self-training, where the former trains an imputation model explicitly using observed values of incomplete data, while the latter uses reliable pseudo values for more accurate training. Self-training employs an imputation model trained with RDI as a teacher model. In the following, we delineate the problem statement, RDI, and self-training.
\subsection{Problem statement}
We denote an original incomplete time-series data as $X=\{x_1,\ldots,x_T\}\in\mathbb{R}^{T\times{D}}$, where each element is a vector $x_t=\{x_t^{1},\ldots,x_t^{D}\}\in\mathbb{R}^{D}$. Since $X$ is an incomplete time-series data, we define a mask, $M\in\{0,1\}^{T\times{D}}$, to indicate the location of the missing values. Each element of the $M$ is expresses as follows:
\begin{align}
    m_t^{d}=
    \begin{cases}
    0, & \text{if}\ x_t^{d} \text{ is a missing value} \\
    1, & \text{otherwise.}
    \end{cases}
\end{align}
We also introduce a complete time-series data, $X_{ideal}\in\mathbb{R}^{T\times{D}}$, which has hypothetical values, considered as ground truth. Then, we can denote $X=X_{ideal} \odot M$, where $\odot$ indicates element-wise multiplication. By using these notations, imputation attempts to find $\theta$ that minimizes the following objective function:
\begin{align}
    \min_{\theta}\mathbb{E}[||X_{ideal}\odot(1-M)-F(X;\theta)\odot(1-M)||_2],
\end{align}
where $F$ is an imputation model with parameter $\theta$. Note that optimizing (2) is a challenging task because $X_{ideal}$ is not existing. Various approaches were taken to solve this problem in the previous imputation studies, and the most common approach is to use the following reconstruction loss:
\begin{align}
    L_{recon}=||X\odot M-F(X;\theta)\odot M||_2,
\end{align}
which is a loss term for reconstructing observed values. (3) can help imputation, but it is different from (2), which learns to fill missing values directly. Therefore, training imputation model using (3) can be called implicit training. In contrast, we propose a new training method using a random drop imputation (RDI) for explicit training.

\subsection{Random drop imputation}
RDI explicitly trains an imputation model by random drop data obtained from randomly removing observed values in the time-series data. Let's denote the random drop data as $\tilde{X}=\{\tilde{x}_1,\ldots,\tilde{x}_T\}\in\mathbb{R}^{T\times{D}}$, where each element is a vector $\tilde{x}_t=\{\tilde{x}_t^{1},\ldots,\tilde{x}_t^{D}\}\in\mathbb{R}^{D}$. We also denote a mask which represents the location of both originally missing values and randomly dropped values, as $\tilde{M}$ of which element $\tilde{m}_t^{d}$ is defined as follows:
\begin{align}
    \tilde{m}_t^{d}=
    \begin{cases}
    0, & \text{if}\ \tilde{x}_t^{d} \text{ is a missing value} \\
    1, & \text{otherwise.}
    \end{cases}
\end{align}

The loss function of RDI, called imputation loss, can be expressed as follows:
\begin{align}
    L_{impute}(\Tilde{X}, \Tilde{M})=&||X\odot(M-\tilde{M})-F(\tilde{X};\theta)\odot(M-\tilde{M})||_2 \nonumber \\& + ||\tilde{X}\odot \tilde{M}-F(\tilde{X};\theta)\odot \tilde{M}||_2,
\end{align}

where the first term of (5) is the core objective function of imputation, which is similar to (2). After the random drop, missing values with ground truth are generated, which can be explicit learning of (2). In contrast, the second term is the reconstruction loss of observed values, which helps stable training. 

Note that as our proposed method, RDI, creates missing values with ground truth, using the first term of (5), we can explicitly train an imputation model. In addition, RDI has one more advantage in terms of data augmentation, in which we can generate various combinations to drop observed values randomly from the original data. By taking this advantage, we generate multiple random drop data from one original data and use them for RDI. Also, to utilize the augmented data, our RDI employs ensemble learning. 

To deal with unstable and over-fitting issues, bootstrap with ensemble learning \cite{schomaker2018bootstrap, tang2017random} was applied to the imputation task. This motivates us to adopt ensemble learning to boost imputation performance. Bootstrap constructs several sets using random sampling and trains models with each set. However, one of the disadvantages of bootstrap is the bias from the random sampling. On the other hand, as RDI creates different sets by augmenting the entire dataset, we can apply ensemble learning without bias.

As shown in Figure 1(a), we generate $N$ different random drop data, denoted as $\tilde{\mathbb{X}}=\{\tilde{X}_1, \ldots, \tilde{X}_N\}$. Then, $N$ imputation models, which are composed of the same structure but trained with different data using the loss function (5) can be denoted as $\mathbb{F}=\{F_1, \ldots, F_N\}$. Each model $F_k$ is trained to generate original data $X$ from random drop data $\tilde{X}_k$. After training, the original data is fed to each pre-trained model $F_k$ for imputation. However, the output values from individual models are all distinctive since differently augmented data are used when training individual models. Therefore, the average of $N$ output values is considered a final output of the ensemble model. 
\begin{figure}
  \centering
  \includegraphics[width=\linewidth]{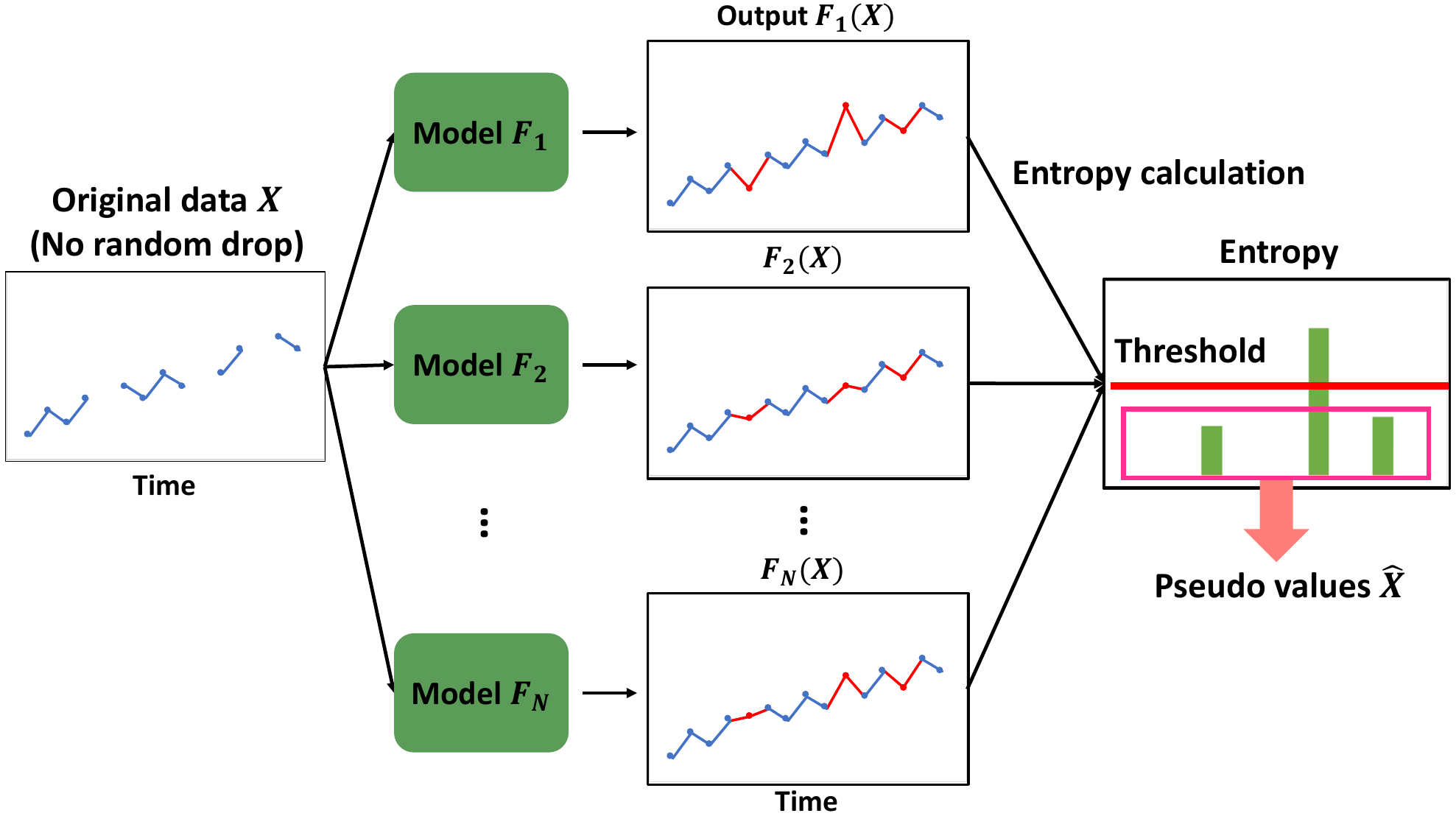}
  \caption{Overall procedure of determining reliable pseudo values. The blue line indicates the observed values, while the red line indicates generated values. In order to choose reliable pseudo values, the entropy of each missing value from $N$ pre-trained models is first calculated. Then, the values with lower entropy than the threshold are selected as pseudo values.}
\end{figure}
\subsection{Random drop imputation with self-training}  
Original incomplete data $X$ contains observed and missing values (without ground truth). RDI only exploits observed values of the original incomplete data. Hence, we apply self-training to exploit the missing values as well as the observed values. We utilize pseudo values of originally missing values for self-training. We feed the original data to the pre-trained models, $F_1, \ldots, F_N$, for generating the pseudo values. Then, the average outputs of the pre-trained models, $\hat{X}=\frac{1}{N}\sum_{k=1}^N F_k(X)$ are considered as pseudo values. The pseudo values are updated periodically every fixed number of epochs, which we call an update epoch. The loss function of each model $F_k$ for the self-training is expressed as follows:
\begin{align}
    L_{self, k}(\hat{X}, \Tilde{X}_k)= &||\hat{X}\odot(1-M)-F_k(\tilde{X}_{k};\theta_{k})\odot(1-M)||_2 \nonumber \\
    &+ ||X\odot M-F_k(\tilde{X}_{k};\theta_{k})\odot M||_2,
\end{align}
where the first term is the loss between missing and pseudo values. The second term is the reconstruction loss of the observed values.

RDIS also works better with an additional filtering rule. We filter out pseudo values that have low confidence. The evaluation of the confidence of pseudo values is different from that of pseudo labels. Usually, most self-training methods filter the pseudo labels using predicted probability from the pre-trained model, but there is no probability in the imputation. Considering this issue, RDIS uses entropy to check whether each pseudo value is reliable or not.

In the case of classification and segmentation, which predict discrete labels, the model's output is a probability of $[0,1]$. Therefore, a deterministic task can easily measure the confidence of a pseudo-label. In contrast, the imputation model predicts continuous values, so the outputs are not a probability. To this end, entropy is employed to calculate the confidence level. We apply ensemble learning to RDI, where $N$ pre-trained models generate $N$ values for each missing value in the original incomplete data. Assuming that these $N$ values follow the Gaussian distribution, the entropy can be calculated as follows:
\begin{align}
    Entropy &= -\int_{-\infty}^{\infty}p(x)\log{p(x)}dx = \ln (\sigma \sqrt{2\pi e}),
\end{align}
The variance of the $N$ values, $\sigma$, is employed to measure the entropy. The output value with low entropy is considered reliable and used as a pseudo value, whereas the output value with high entropy is considered uncertain and discarded. The threshold value that distinguishes reliable/unreliable entropy is determined empirically through experiments. The overall procedure of selecting pseudo values is shown in Figure 2. For calculation convenience, we set the pseudo value threshold as the variance of the $N$ values. The pseudo-code of the proposed RDIS is illustrated in Alg.\ref{alg:one}. 

\RestyleAlgo{ruled}
\SetKwInput{KwInput}{Input}
\SetKwInput{KwOutput}{Output}
\SetKwInput{KwParameters}{Parameters}
\begin{algorithm}
\caption{Training strategy of RDIS}\label{alg:one}
    \KwInput{Incomplete time series data $X$, mask $M$, pre-trained imputation models $F_1(X;\theta_1), \ldots, F_N(X;\theta_N)$ from RDI}
    \KwParameters{$N, \tau, numEpochs, numUpdateEpochs$}
    \KwOutput{Trained N imputation models $F_1, \ldots, F_N$}
    $e \gets 0$
    
    \While{$e < numEpochs$}{
        \If{$e\ \% \ numUpdateEpochs == 0$}{
            Empty list containing pseudo values: $L \gets []$\\
            \For{i=1, \ldots, N}{
            $L \gets  Concatenate(L; F_i(X, \theta_i))$ // final dimension of $L: (N,T,D)$ \\ 
            }
            \For{t=1, \ldots, T}{
                \For{d=1, \ldots, D}{
                    Compute variance $\sigma$ of time step $t$ and dimension $d$ using data in $L$\\
                    \If{$\sigma < \tau$}{
                        $\hat{x}_t^d \gets \frac{1}{N}\sum_{i=1}^N L_{i, t, d}$ // generate pseudo values\\
                    }
                }
            }
        }
        \For{i=1, \ldots, N}{
            $F_i \gets L_{self, i}(\hat{X}, \Tilde{X}_i)$ // update each model using pseudo values\\
        }
        $e \gets e+1$\\
    }
\end{algorithm}

\begin{table*}[t]
\centering
\begin{tabular}{@{}c|c|cccccccc@{}}
\toprule
\multirow{2}{*}{Dataset}      & \multirow{2}{*}{Method} & \multicolumn{8}{c}{missing rate}                                                                                                              \\
                              &                           & 10\%            & 20\%            & 30\%            & 40\%            & 50\%            & 60\%            & 70\%            & 80\%            \\ \midrule
\multirow{12}{*}{Air Quality} & Forward                   & .2472          & .2638          & .2707          & .2694          & .2747          & .3264          & .3888          & .4432          \\
                              & Backward                  & .1849          & .2137          & .2847          & .2832          & .2799          & .3333          & .3624          & .4445          \\
                              & MICE                      & .3405          & .4135          & .3888          & .4235          & .4309          & .5663          & -               & -               \\
                              & KNN                       & .1520           & .2323          & .2847          & .3501          & .3930           & .4327          & .4843          & .5775          \\ \cmidrule(l){2-10} 
                              & GRU                       & .3137          & .3617          & .3757          & .3735          & .3834          & .4331          & .4523          & .4962          \\
                              & Bi-GRU                    & .2706          & .3136          & .3336          & .3287          & .3363          & .3902          & .4029          & .4445          \\
                              & M-RNN                      & .1583          & .2127          & .2342          & .2324          & .2417          & .2919          & .3248          & .4038          \\
                              & BRITS                     & .1659          & .2076          & .2212          & .2088          & .2141          & .2660          & .2885          & .3421          \\
                               \cmidrule(l){2-10} 
                              & RDIS (GRU)                     & .1854 & .2385 & .2600 & .2557 & .2629 & .3075 & .3276 & .3826 \\
                              & RDIS (Bi-GRU)                  & \textbf{.1409} & \textbf{.1807} & \textbf{.2008} & \textbf{.1977} & \textbf{.2041} & \textbf{.2528} & \textbf{.2668} & \textbf{.3178} \\ \midrule
\multirow{12}{*}{Gas Sensor}  & Forward                   & .0834          & .0987          & .1181          & .1460          & .1845          & .2398          & .3215          & .4739          \\
                              & Backward                  & .0838          & .1016          & .1267          & .1522          & .1894          & .2449          & .3310          & .5501          \\
                              & MICE                      & .0736          & .0822          & .1090          & .1468          & .2147          & -          & -               & -               \\
                              & KNN                       & .0328           & .0356          & .0388          & .1273          & .2423           & .3771          & .4921          & .8009          \\ \cmidrule(l){2-10} 
                              & GRU                       & .0727          & .0838          & .0940          & .1127          & .1349          & .1647          & .1998          & .3223          \\
                              & Bi-GRU                    & .0657          & .0766          & .0862          & .1024          & .1235          & .1451          & .1802          & .2903          \\
                              & M-RNN                      & .0327          & .0400          & .0378          & .0388          & .0405          & .0427          & .0435          & .1023          \\
                              & BRITS                     & \textbf{.0210}          & \textbf{.0226}          & \textbf{.0233}          & .0279          & .0338          & .0406          & .0518          & .1595          \\
                               \cmidrule(l){2-10} 
                              & RDIS (GRU)                      & .0239 & .0255 & .0276 & .0311 & .0373 & .0434 & .0506 & .1070 \\
                              & RDIS (Bi-GRU)                   & .0287 & \textbf{.0226} &  .0241 & \textbf{.0251} & \textbf{.0277} & \textbf{.0321} & \textbf{.0350} & \textbf{.0837} \\ \bottomrule
\end{tabular}
\caption{The MSE comparison of imputation methods under different missing rates on both the air quality dataset and the gas sensor dataset}
\end{table*}

\section{Experiment}
In this section, we applied RDIS to both GRU and bidirectional GRU to test the effectiveness of RDIS. For comparison, we employed the previous machine-learning methods and deep-learning-based methods. We evaluated the performance on two real-world time series datasets: air quality and gas sensor datasets.

\subsection{Dataset Description}
\subsubsection{Air Quality Data}
Beijing Multi-Site Air-Quality dataset \cite{airquality} from the UCI machine learning repository, abbreviated as air quality dataset, was used in the experiment. The air quality dataset is air pollutant data from 12 nationally-controlled air-quality monitoring sites. This measurement was collected hourly from 03/01/2013 to 02/28/2017. The dataset consists of 12 features. Among the features, we excluded the direction of the wind because it is implausible to represent the value numerically. We randomly selected 48 consecutive time steps to generate one time series.

\subsubsection{Gas Sensor Data}
Gas Sensor Array Temperature Modulation dataset \cite{gassensor} from UCI machine learning repository, abbreviated as gas sensor dataset, was also used in the experiment. The gas sensor dataset comprises 19 features, including humidity, CO (ppm), and temperature. We randomly select 48 consecutive time steps to generate one time series.

\subsection{Baseline Methods}
For the purpose of imputation performance comparison, we used the following baselines:
\begin{itemize}
\item $\textbf{Forward/Backward:}$ The missing values are simply imputed with the last forward/backward observed values.
\item $\textbf{MICE:}$ The incomplete data are estimated with a low-rank approximation, and the missing values are filled with the singular vectors \cite{royston2011multiple}.
\item $\textbf{KNN:}$ $\textit{k}$-nearest neighbors (KNN) \cite{friedman2001elements} imputes the missing values with the weighted average of the $k$ nearest sequences.
\item $\textbf{GRU/Bi-GRU:}$ The uni/bidirectional autoregressive Gated Recurrent Unit (GRU) \cite{chung2014empirical, cho2014learning} model is for imputing missing values. The hidden states are fed to a fully connected layer to obtain the output.
\item $\textbf{M-RNN:}$ M-RNN \cite{yoon2017multi} uses a bidirectional RNN. M-RNN estimates the missing values using the hidden states across streams in addition to within streams.
\item $\textbf{BRITS:}$ BRITS \cite{cao2018brits} imputes missing values using a bidirectional RNN and considers correlation among different missing values.
\end{itemize}
We used Impyute\footnote{https://github.com/eltonlaw/impyute.} to implement MICE and fancyimpute\footnote{https://github.com/iskandr/fancyimpute.} to implement KNN.

\begin{figure*}[t]
    \centering
    \includegraphics[width=\linewidth]{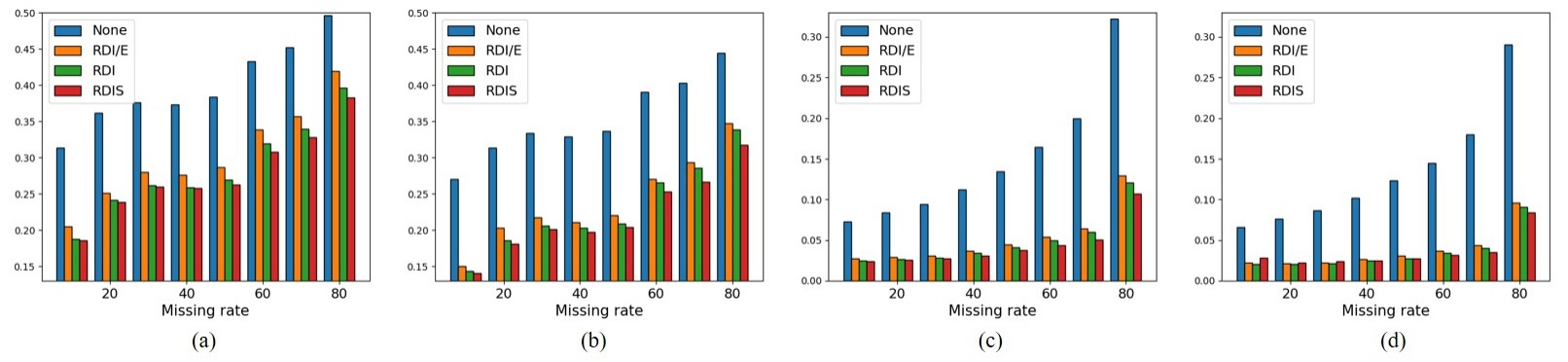}
    \caption{The MSE comparison of GRU and Bi-GRU grafted with None, RDI/E, RDI, and RDIS, respectively, on the air quality dataset and the gas sensor dataset. (a) Air quality (GRU), (b) air quality (Bi-GRU), (c) gas sensor (GRU), (d) gas sensor (Bi-GRU).}
\end{figure*}

\subsection{Experimental Settings}
We normalized each dataset to have zero mean and unit variance. For each dataset, we selected 50\% of the data for training, 25\% for validation, and the remaining 25\% for testing in sequential order. The following model training parameters were used in all experiments for both datasets. The input batch size of 128 and the Adam optimizer were used for training. The learning rate and the number of epochs were set to $5\mathrm{e}{-4}$ and 2000, respectively. The dimension of the hidden unit was set to 100 for all RNN-based methods. The hyper-parameters (random drop rate, the threshold of pseudo value, and the pseudo value update epoch) were tuned empirically. Each experiment was evaluated five times, and the imputation performance was evaluated in terms of mean square error (MSE). We evaluated the imputation performance under different missing rates to show that RDIS is robust to various missing rates. Also, original missing values cannot measure MSE because there is no ground truth. In this light, we randomly discarded $p$\% of the observed data and used them as the ground truth and selected $p$ from $\{10,\ 20,\ 30,\ 40,\ 50,\ 60,\ 70,\ 80\}$. In the test stage, we use the average of the $N$ models as the final output.

\subsection{Performance Comparisons Studies}
To verify the effectiveness of our RDIS on imputation tasks, we grafted RDIS to GRU and Bi-GRU, of which performances were compared with other baseline methods. Table 1 shows the MSE on the air quality and the gas sensor dataset. On both datasets, we used 8 models in RDIS for ensemble learning. For the air quality dataset, the random drop rate, the pseudo value threshold, and the pseudo value update epoch were set to 30\%, 0.03, and 400, respectively. Likewise, for the gas sensor dataset, we used 20\%, 0.015, and 400, respectively. Since BRITS and M-RNN were specialized for time series imputation, they achieved low imputation errors at all missing rates. However, RDIS with Bi-GRU achieved the lowest error under most conditions. RDIS with Bi-GRU obtained competitive performance over the state-of-the-art methods on the air quality dataset and the gas sensor dataset.

\subsection{Ablation Studies}
To analyze the contribution of RDIS, we grafted it onto GRU and Bi-GRU, and we conducted experiments on four different settings: the model without RDIS (None), the model with RDI but without ensemble learning (RDI/E), the model with RDI, and the model with RDIS. We used the air quality and gas sensor dataset under different missing rates, and the same hyper-parameters employed in Table 1 were used. Figure 3 exhibits the effects of RDI/E, RDI, and RDIS. First, all our methods dramatically improved in all cases. It implies that learning with our methods led to the proper training of the imputation models. Also, RDI showed better performance than RDI/E. It demonstrates the capability of ensemble learning. Our final observation is that RDIS generally showed the best performance, and the higher the missing rate, the higher the performance improvement. As the missing rate increases, an imputation model quickly overfits due to a lack of data. Nevertheless, RDIS can prevent overfitting by replacing missing values with reliable pseudo values.

\subsection{Hyper-parameter Studies}
In this section, we analyze hyper-parameters (random drop rate, threshold of pseudo value, and pseudo value update epoch), which control the imputation performance.
\begin{table}
\centering
\begin{tabular}{@{}cccccc@{}}
\toprule
Dataset$\backslash$RDR & \multicolumn{1}{c}{10\%} & \multicolumn{1}{c}{20\%} & \multicolumn{1}{c}{30\%} & \multicolumn{1}{c}{40\%} & \multicolumn{1}{c}{50\%} \\ \midrule
Air Quality                       & .2501                   & .2124                   & \textbf{.2091}          & .2187 & .2615 \\
Gas sensor                    & .0348                   & \textbf{.0278}                   & .0352          & .0473 & .0714\\ \bottomrule
\end{tabular}
\caption{Analysis of random drop rate on RDI}
\end{table}
\begin{figure}
    \centering
    \includegraphics[width=\linewidth]{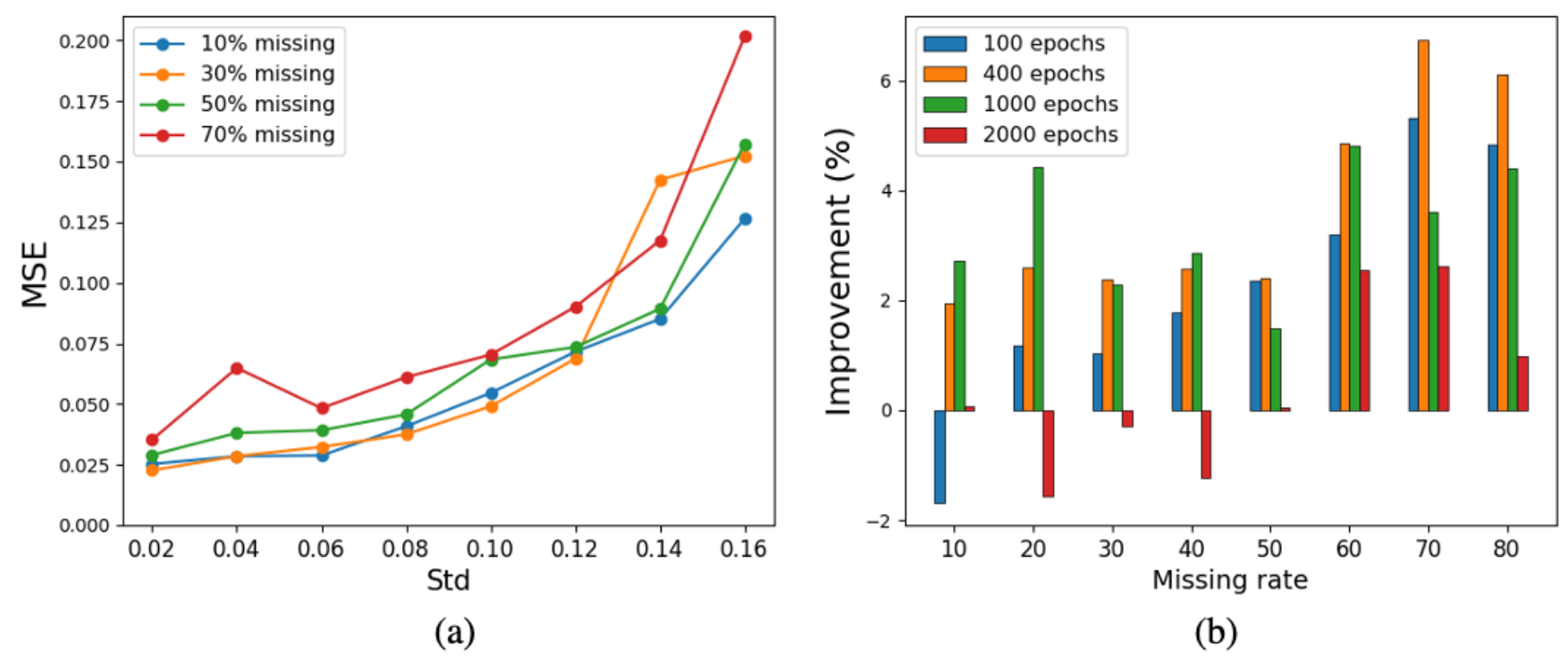}
    \caption{Pseudo value and update epoch analyses. (a) Pseudo value's accuracy per standard deviation over different missing rates, (b) the ratio of MSE reduction (\%) of RDIS based on RDI for different update epochs.}
\end{figure}
\subsubsection{Random Drop Rate Analysis}
We analyzed the effect of the random drop rate (RDR) using two datasets. For this analysis, we used Bi-GRU with RDI, and each dataset was tested for a missing rate of 50\%. Table 2 summarizes experimental results on the imputation performances for RDRs of 10\%, 20\%, 30\%, 40\%, and 50\%. In the air quality dataset, the best performance was obtained when RDR was 30\%, and in the gas sensor dataset, the best performance was obtained at 20\%.
\subsubsection{Pseudo Value Analysis} 
In order to select reliable pseudo values, we measured the entropy of each pseudo value. To verify our proposed reliability measurement method, we computed pseudo value's accuracy according to entropy. The entropy is proportional to $\ln(\sigma)$, as shown in (7), so we report pseudo value's accuracy according to the variance for convenience of calculation. We used the air quality dataset under different missing rates for this experiment. We also used Bi-GRU with RDIS, and we fixed an update epoch as 400. Figure 4 (a) shows that the pseudo value's accuracy declines when the variance increases. It means that pseudo values with low variance are reliable ones, which verifies the effectiveness of our proposed pseudo value selection method.

\subsubsection{Update Epoch Analysis}
We updated pseudo values periodically at a fixed epoch. To evaluate the effect of pseudo value update epoch on RDIS, we experimented with update epochs of 100, 400, 1000, and 2000, respectively. In this experiment, Bi-GRU with RDIS was employed on the air quality dataset under different missing rates, and a variance of 0.03 was used as a threshold for selecting the pseudo value. We report the rate of reduction in MSE of RDIS compared to that of RDI. Figure 4 (b) shows the result of this experiment. It is claimed that frequent updates of pseudo values can negatively affect imputation performance. The reason is that the direction to which the model should converge changes too often. On the other hand, too late update of pseudo values can also cause a negative effect. This is because the pseudo values created by the model with improved performance through self-training are more accurate.

\section{Conclusion}
This paper proposed a novel explicit training method, RDIS, consisting of RDI and self-training for imputing incomplete time-series data. Proposed RDIS learned imputation explicitly via random drop data augmented from the original data. In order to fully make use of random drop data, RDI adopted ensemble learning, which significantly increased imputation performance. In addition, self-training was incorporated into RDI to design RDIS, which was first introduced in imputation. We evaluated the performance of the air quality and the gas sensor dataset. Experimental results show that the bidirectional GRU with RDIS achieved state-of-the-art results with a significant margin. For future work, we test RDIS to tabular data imputation. Also, adding loss for high uncertainty level pseudo values shall be investigated.

\bibliographystyle{./IEEEtran}
\bibliography{./ref}

\vspace{12pt}

\end{document}